\documentclass{bmvc2k}
\usepackage{amssymb}
\usepackage{enumitem}
\usepackage{amsmath}
\usepackage{amssymb}
\usepackage{amsmath}
\usepackage{enumitem}
\usepackage{graphicx}
\usepackage{color, colortbl}
\usepackage{tabularx}
\usepackage{tikz}
\usepackage{float}
\usepackage{threeparttable}

\newcommand{\figref}[1]{Fig.~\ref{#1}}
\newcommand{\tabref}[1]{Table~\ref{#1}}

\newcommand{\sref}[1]{Sec.~\ref{#1}}

\title{Sequential Feature Filtering Classifier}

\addauthor{Minseok Seo\textsuperscript{*}}{msseok96@gmail.com}{1}
\addauthor{Jaemin Lee\textsuperscript{*}}{j911.public@gmail.com}{1}
\addauthor{Jongchan Park}{jcpark@lunit.io}{2}
\addauthor{Donggeol Choi}{dgchoi@hanbat.ac.kr}{1}
\addinstitution{
Hanbat National University, \\
Daejeon, Korea
}
\addinstitution{
 Lunit Inc.\\
 Seoul, Korea
}

\runninghead{Seo, Lee, Park, Choi}{Sequential Feature Filtering Classifier}

\newcommand{\jc}[1]{\textcolor{black}{#1}}
\newcommand{\jctwo}[1]{\textcolor{black}{#1}}
\newcommand{\jcthree}[1]{\textcolor{black}{#1}}
\begin{document}

\maketitle

\begin{abstract}
We propose Sequential Feature Filtering Classifier (FFC), a simple but effective classifier for convolutional neural networks (CNNs).
With sequential LayerNorm and ReLU, FFC zeroes out low-activation units and preserves high-activation units.
\jcthree{
The sequential feature filtering process generates multiple features, which are fed into a shared classifier for multiple outputs.}
FFC can be applied to any CNNs with a classifier, and significantly improves performances with negligible overhead.
We extensively validate the efficacy of FFC on various tasks: ImageNet-1K classification, MS COCO detection, Cityscapes segmentation, and HMDB51 action recognition. 
Moreover, we empirically show that FFC can further improve performances upon other techniques, including attention modules and augmentation techniques. The code and models will be publicly available.
\end{abstract}

\section{Introduction}
\label{sec:intro}
Deep convolutional neural networks(CNNs) have made remarkable advances in multiple computer vision tasks such as image classification~\cite{deng2009imagenet}, object detection~\cite{lin2014microsoft}, semantic segmentation~\cite{cordts2016cityscapes}, and action recognition~\cite{kuehne2011hmdb}.
Architecture design is one of the main research topics for CNN.
A number of backbone architectures have been proposed to push the state-of-the-art, such as AlexNet~\cite{krizhevsky2012imagenet}, VGG~\cite{simonyan2014very}, Inception~\cite{szegedy2015going}, ResNet~\cite{he2016deep}, ResNeXt~\cite{xie2017aggregated}, NASNet~\cite{zoph2018learning}, EfficientNet~\cite{tan2019efficientnet}, etc. 
To improve the performance of a given task, the simplest option is to change the CNN backbone to bigger or stronger ones. However, it may be limited by memory or computational constraints, or by overfitting phenomena.
Apart from CNN architectures, simple techniques with little or no overheads have also been proposed to improve the overall generalization performances of CNNs.
Data augmentation~\cite{devries2017improved, ghiasi2018dropblock, zhang2017mixup, yun2019cutmix} is a widely used technique for regularization. Optimization methods such as~\cite{hinton2015distilling, loshchilov2016sgdr} are also preferred for the same reason. Adding more data augmentations and changing optimization techniques require no architecture changes, and no test-time overheads, but may require longer training time. Self-attention methods~\cite{hu2018squeeze, park2018bam, woo2018cbam, hu2018gather, wang2018non, cao2019gcnet, huang2019ccnet, lee2019srm} can also be used to improve performances with little overheads.  They can be attached to any CNN architectures and trained end-to-end without bells and whistles. These techniques have shown consistent improvements in a wide range of tasks with various backbone architectures. In real world scenarios, utilizing such off-the-shelf techniques is crucial to achieve the best performance out of a given dataset and given hardware constraint.

\begin{figure*}[!htb]
  \centering
  \includegraphics[width=\linewidth]{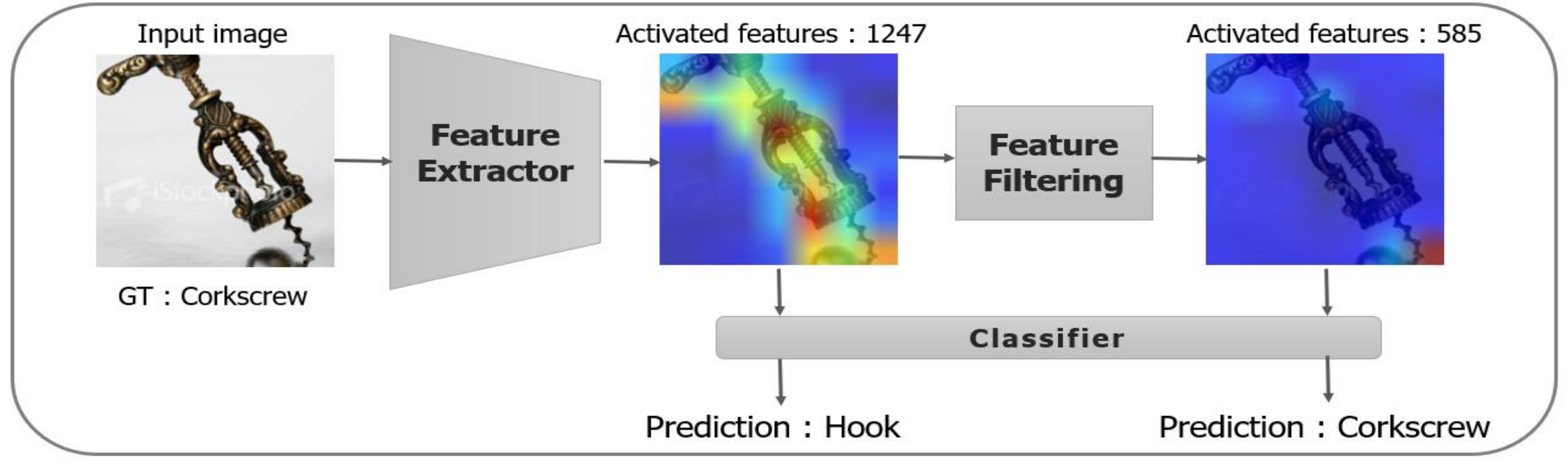}
   \vspace{1mm}
  \caption{\textbf{Predicted results by adjusting the activation unit using FFC.} The image is a visualization of whether or not the activation function was controlled by FFC using Grad-CAM~\cite{selvaraju2017grad}(Model attached to FFC on ResNet-50 in ImageNet-1K and trained. The number of features that are finally extracted from ResNet-50 and entered into the classifier is 2048. )}
  \label{fig:intuition}
  \vspace{-4mm}
\end{figure*}

In this work, we propose a simple and effective off-the-shelf classifier, named Sequential Feature Filtering Classifier (FFC), that can be used in any CNN architectures and in various target tasks.
Our intuition is that the activated features of samples in each class may differ due to high intra-class variations, and we can exploit the given feature by emphasizing high-activation features and suppressing low-activation features.
\figref{fig:intuition} shows the different results by adjusting the number of activated features: the number of activated features, the preliminary predictions, and the associated Grad-CAM sample. As shown in \figref{fig:intuition}, if the classifier makes predictions by looking at the entire feature, it is incorrectly predicted as "Hook". This is because "Hook" feature is included in the sample image feature. However, if only the highly activated features are entered as a classifier through feature filtering, we can see only the features with high activation rather than the overall features of the sample image, so we can see that CNN have accurately predicted "Corkscrew".



FFC consists of sequential LayerNorm layer~\cite{ba2016layer} and ReLU~\cite{nair2010rectified}, and a single shared fully-connected layer. 
The overall structure of FFC is shown in \figref{fig:module}. Given a feature vector, multi-level features are calculated with LayerNorm and ReLU in a sequantial manner. The multi-level features are then ensembled by a simple combination rule and fed into the shared classifier. The multiple outputs are combined as the final prediction. Simply put, FFC exploits different versions of a single feature generated by sequential filtering.
Compared to a simple fully-connected classifier, FFC has negligible parameter and computational overheads. For example, when the backbone is ResNet-50~\cite{he2016deep}, FFC has a negligible parameter overhead of 0.012M, and a negligible computational overhead of 0.014 GFLOPS. Due to its light-weightness and simplicity, FFC can be widely used in any CNN architectures and in any recognition tasks. 

We have extensively validated the efficacy of FFC in various architectures and tasks. We evaluated FFC in Imagenet-1K classification, MS COCO object detection, Cityscapes segmentation, and HMDB51 action recognition. FFC has shown consistent and significant improvement over every architectures and tasks. We further show that FFC can improve the performance upon off-the-shelf techniques, such as attention modules and bag-of-tricks, empirically showing that FFC has complementary effects.

\smallskip\noindent\textbf{Contribution.} Our main contribution is three-fold.
\begin{enumerate}[topsep=0pt,itemsep=0pt]
\item \jc{We propose a simple classifier, named FFC, that sequentially filters feature vectors and combines multiple outputs with negligible overheads.}
\item \jc{We empirically show the complementary effects of FFC against other off-the-shelf techniques such as fine-tuning, attention modules, data augmentations, and ensemble.}
\item We verify the effectiveness of FFC throughout extensive experiments with various baseline architectures on multiple benchmarks (ImageNet-1K, MS COCO, Cityscapes, HMDB51).
\end{enumerate}

\section{Related Work}\label{sec:related}

In this section, we investigated dropout and cascade architecture, which are similar in concept to FFC. Also, we investigated technologies such as attention modules, MixUp, knowledge distillation, and SGDR, which improve performance regardless of CNN architecture. \\ \\
{\bf Dropout.} Dropout is introduced by Hinton et al.~\cite{hinton2012improving} and Sivrastava et al.~\cite{srivastava2014dropout}. Their key idea is to randomly drop units (along with their connections) from the neural network during training. 
Dropout makes the network less overfitted to the training data by preventing feature units from excessive co-adaptation.
Since then, many works have been proposed~\cite{wan2013regularization, li2016improved, ba2013adaptive}, all sharing the philosophy of Dropout.
FFC is similar to Dropout in that it zeros out units of extracted features. 
However, while Dropout randomly zeroes out units, FFC sequentially zeroes out smallest non-zero units with LayerNorm and ReLU.
Eventually,  FFC uses sequentially extracted units, and sequentially generates various levels of features from the same feature and inputs it to the shared classifier.
For this reason, multiple levels of multi-output come out, and it works during testing because it is used by voting when testing. \\ \\
{\bf Cascade Classifier.}
Cascade R-CNN~\cite{cai2018cascade} uses sequential classifiers where each classifier takes the refined bounding box from the previous classifier. The bounding box is sequentially refined, the prediction is expected to be more and more accurate.
FFC and Cascade R-CNN are similar in that both augments the classifier part of a network and have multiple outputs. Cascade R-CNN requires multiple classifiers for multiple outputs, but FFC uses a shared classifier, so much less parameter overheads are required.
Also, while Cascade R-CNN is explicitly designed and validated for the object detection task only, FFC has a generally applicable design and is validated for multiple recognition tasks. \\ \\ 
{\bf Bag of Tricks.}
There are simple, off-the-shelf techniques to further boost the performance of a given model.
He \textit{et al.}~\cite{he2019bag} investigated the augmentation and optimization techniques, and called them ``Bag of Tricks''. The ``Bag of Tricks'' consists of techniques such as MixUp~\cite{zhang2017mixup}, knowledge distillation~\cite{hinton2015distilling}, and cosine analysis learning rate decay~\cite{loshchilov2016sgdr} mentioned earlier.
With these \textit{tricks}, the top-1 accuracy of ResNet-50 in ImageNet is increased from the baseline performance 75.3\% to 79.29\%. The result shows that such \textit{tricks} are crucial to achieve the best performance out of a given architecture.

Like the aforementioned \textit{tricks}, FFC is a general module that can be used for various architecture and tasks. It has a very simple design and negligible overheads, but brings significant performance improvements.
As an extensive validation of FFC, we show that FFC can bring further improvements upon the \textit{tricks}. \\ \\
{\bf Attention method.} The attention method~\cite{hu2018squeeze, park2018bam, woo2018cbam, hu2018gather, wang2018non, cao2019gcnet, huang2019ccnet, lee2019srm} is a well-known technique that improves the performance of CNN with less parameters and computation. Even a simple implementation can be attached to most CNN architecture, making it one of the preferred technologies. Squeeze-and-Excitation (SE) improved the performance of CNN through a channel-wise recalibration operator, and the Convolutional block attention module (CBAM) achieved a further improved performance by adding a spatial attention module. Lastly, the recently proposed Style-based Recalibration Module (SRM) improved the performance of CNN by proposing a channel-independent style integration method utilizing style pooling. These technologies are located between feature maps and feature maps in CNN's feature extractor. However, our FFC is located just before the final feature enters the classifier. Therefore FFC can be used with attention modules and has been experimentally verified to be complementary.

\section{Sequential Feature Filtering Classifier}
\label{sec:FOC}
\begin{figure*}[!t]
  \centering
  \includegraphics[width=\linewidth]{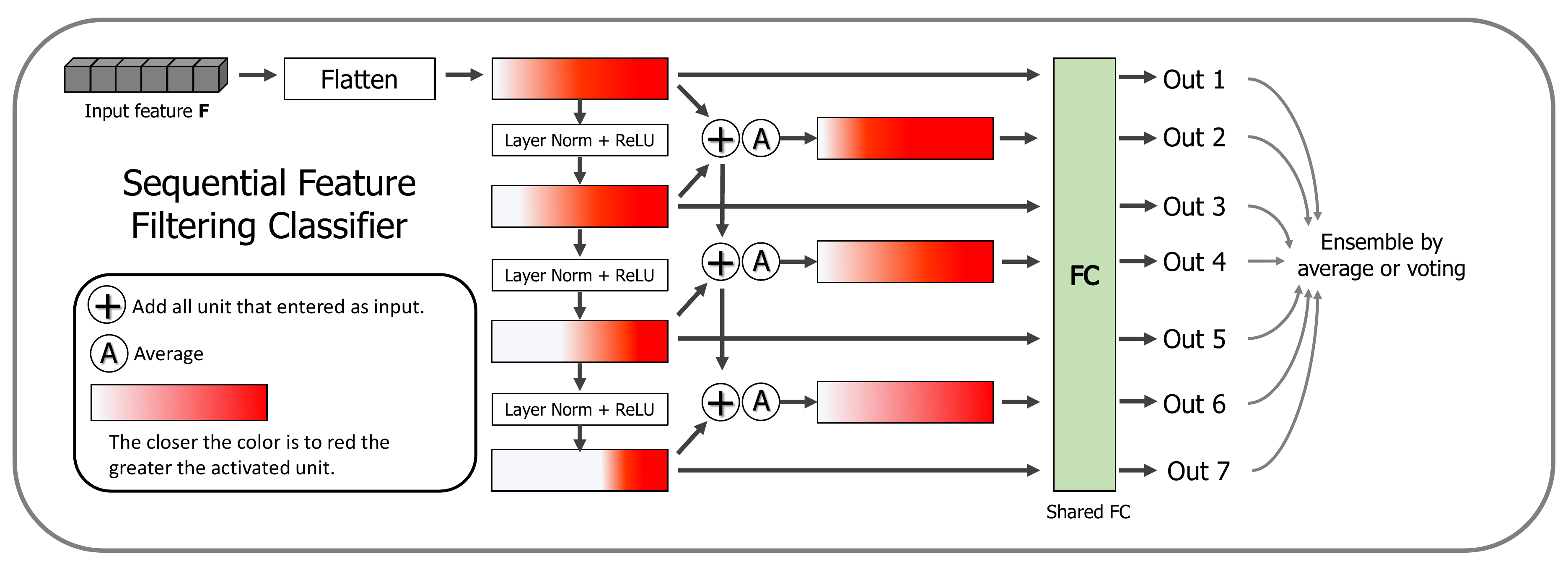}
    \vspace{2mm}

  \caption{\textbf{FFC integrated with a general CNN architecture.} FFC is placed after the final feature extracted by the feature extractor. FFC receives a flattened feature as input, highlights the feature with a large value activated through sequential LayerNorm, ReLU, and enters it into the shared FC. FC receives features of various scales as inputs and performs various outputs. Finally, we ensemble the outputs from the shared FC to make final predictions.}
  \label{fig:module}
  \vspace{-4mm}
\end{figure*}

\jc{In this section, we describe the algorithm of FFC. We will first review a well-known normalization layer, LayerNorm~\cite{ba2016layer}, and then describe how FFC is designed. We choose image classification with 2D inputs as an example, but please note that FFC can be easily extended to detection, segmentation, and action recognition where the inputs are 3-dimensional.}\\ \\ 
{\bf Layer Normalization.} In the case of 2D images, \textit{i} = ($\textit{i}_D$, $\textit{i}_C$, $\textit{i}_H$, $\textit{i}_W$) is a 4D vector indexing the features in (N, C, H, W) order, where N is the batch \jc{size}, C is the channel \jc{size}, and H and W are \jc{the height and width}. 
\jc{In image classification, the final feature vector is calculated by a global average pooling with flatten on the intermediate feature map. The normalization statistics are calculated as per-sample basis.
For each sample in the mini-batch}, $\textit{C} = (\textit{c}_1, \textit{c}_2,..., \textit{c}_T)$ is an input feature, and T is the number of units. LayerNorm in FFC re-centers and re-scales input \textit{C} as
\begin{align}
\mu = \frac{1}{T} \sum\limits_{i=1}^{T} \textit{c}_i,\hspace{0.225cm} \sigma = \sqrt{\frac{1}{T} \sum\limits_{i=1}^{T} (\textit{c}_i-\mu)^{2}},\hspace{0.225cm} \textnormal{LN}(\textnormal{F})=\frac{\textnormal{F}-\mu}{\sigma}
\end{align}
$\mu$ and $\sigma$ are the mean and standard deviation of input, respectively.

When layer normalization is applied, values smaller than the average become negative, and \jc{will be zeroed-out by the following ReLU in FFC}.
Therefore, by using layer normalization and ReLU, the number of active units can be dynamically adjusted \jc{on a per-sample basis}. \\ \\
{\bf Sequential Feature Filtering Classifier.} The specific structure of FFC is shown in \figref{fig:module}. If there is an input feature map $\textit{F}\in{\mathbb{R}^{(N \times C \times H \times W)}}$, use avgpool to make the size of the input feature map $\textit{F}\in{\mathbb{R}^{(N \times C \times 1 \times 1)}}$, then flatten the feature map to $\textit{F}\in{\mathbb{R}^{(N \times C)}}$. In general plane network, input $\textit{F} \in{\mathbb{R}^{(N \times C)}}$ into Classifier($C \times Number Of Classes$) and predict the output $\textit{O} \in{\mathbb{R}^{(N \times Number Of Classes}}$). However, our FFC uses $\textit{F}\in{\mathbb{R}^{(N \times C)}}$ as a Sequential Layer Norm and ReLU to zero out units with relatively low activation values compute by:
\begin{align}
\textnormal{F}_1 = \textnormal{ReLU}(\textnormal{LN}(\textnormal{Flatten}(\textnormal{AvgPool}(\textnormal{F})))),\hspace{0.225cm} \textnormal{F}_i = \textnormal{ReLU}(\textnormal{LN}(\textnormal{F}_{i-1}))
\end{align}
In summary, since the sequential Layer Norm re-centers and re-scales the input F, which has only channels, a relatively small unit becomes a negative value, and passes through the ReLU and zeroes out.

FFC has different behavior during training and testing. In common, the shared classifier produces multiple outputs with multiple features from sequential feature filtering.
During training, multiple outputs are ensembled via averaging; during testing, multiple outputs are ensembled via voting.\\

\noindent{\bf Voting rule.} The mean softmax value shown in \tabref{table:ablation} is the average value of the probability of each output predict the correct answer in the ImageNet-1K validation set.  As shown in \tabref{table:ablation}, the Out-1's Top-1 accuracy is not as low as 72\%, but the average softmax value is very low at 0.0017\%, so if we ensemble averaging all the outputs, the meaning of out1 disappears. That's why we have adopted voting, a method that gives equal weight to all outputs. However, if the result of the vote is a tie, the output with the highest confidence value is adopted as the correct predict.

\begin{table}[!hb]
\centering
\resizebox{0.9\textwidth}{!}{%
\begin{tabular}{ l|c|c|c|c}
\hline
Architecture            &  Mean Activated Unit & Sequence & Mean confidence & Top-1\\
\hline
\hline
ResNet-50~\cite{he2016deep} & -   &-     & - &75.85\%\\
ResNet-50~\cite{he2016deep}+FFC(Out1) &1576   &0 &  0.17\%   &72.68\%\\
ResNet-50~\cite{he2016deep}+FFC(Out3) &1176   &1 &  93.81\%    &75.60\%\\
ResNet-50~\cite{he2016deep}+FFC(Out5) &985   &2 &  92.28\%    &76.46\%\\
ResNet-50~\cite{he2016deep}+FFC(Out7) &692   &3 &  88.72\%     &76.06\%\\
ResNet-50~\cite{he2016deep}+FFC(AVG)  &-     &-   & - &76.51\%\\
ResNet-50~\cite{he2016deep}+FFC(VOTE)  &-     &- & -&\textbf{76.80}\%\\
\hline
\end{tabular}
}
\begin{tablenotes}
\scriptsize
\item\hspace*{\fill}\textbf{*} all results are reproduced in the PyTorch framework. 
\end{tablenotes}
\vspace{3mm}
\caption{\textbf{Comparison of different output with FFC.} In our FFC experiment, the vote method achieved the best performance. In addition, FFC showed an accuracy of \textbf{1.1\%} higher than the base with only the parameter \textbf{0.012M} for the layer norm.(Validation in ImageNet-1K)}
\label{table:ablation}
\vspace{-4mm}
\end{table}

Our FFC uses all the outputs (Out1-7) in training and testing as shown in \figref{fig:module}. However, Out2, Out4, and Out6 do not have a zero out unit because the features are averaging. (There is only a change in the scale of the unit.) Therefore, in order to analyze the accuracy according to the number of active units, Out2, Out4, Out6 in \tabref{table:ablation} excluded.

\begin{figure*}[!tb]
  \centering
  \includegraphics[width=\linewidth]{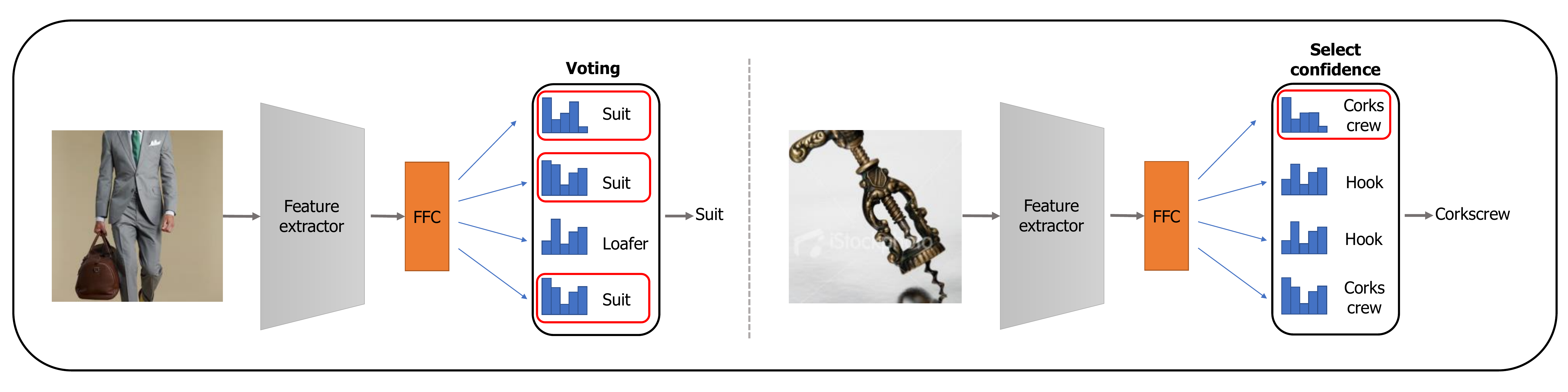}
  \vspace{-2mm}
  \caption{\textbf{FFC voting rule.} In FFC, training time and testing time operate differently. When testing, the final prediction is selected by voting multiple outputs from FFC as shown on the left. However, as shown in the figure on the right, if there is a tie, the softmax value with the highest value (the most confident) is selected and used as the final prediction.}
  \label{fig:voting_rule}
  \vspace{-5mm}
\end{figure*}
\section{Experiments}
\label{sec:Experiments}
We evaluate FFC on the standard benchmarks: ImageNet-1K image classification, MS COCO object detection, Cityscapes semantic segmentation, HMDB51 action recognition in video, visualization using Grad-CAM. In addition, experiments were conducted in various attention modules and bag-of-tricks to verify the complementarity of FFC. We describe in detail our experiment settings for each session and briefly mention the benefits of FFC.
\subsection{Image Classification on ImageNet-1K} \label{sec:ImageNet}
The ImageNet-1K dataset~\cite{deng2009imagenet} is a widely used benchmark for image classification, and it consists of 1.2 million images for training and 50,000 for validation. 
\jc{In this task, we applied FFC on several variants of ResNet and ResNeXt.}
We followed implementation details in ResNet~\cite{he2016deep} and used single-crop evaluation.

\begin{table}[!htbp]
\small
\setlength{\tabcolsep}{3pt}
\begin{center}
\resizebox{1.0\textwidth}{!}{
\begin{tabular}{ l|c|c|c||l|c|c|c}
\hline
Architecture & Params & GFLOPs & Top-1 & Architecture & Params & GFLOPs & Top-1\\
\hline
\hline
ResNet-18~\cite{he2016deep} &11.69M &1.81 &70.40\%
&ResNet-101~\cite{he2016deep} &44.55M &7.57 &76.62\% \\
ResNet-18~\cite{he2016deep}+FFC &11.69M &1.82 &\textbf{70.70\%}
&ResNet-101~\cite{he2016deep}+FFC &44.56M &7.58 &\textbf{77.91\%}\\
ResNet-34~\cite{he2016deep} &21.8M &3.66 &73.31\% 
&ResNeXt-50~\cite{xie2017aggregated}(32x4d) &25.03M &3.77 &77.19\% \\
ResNet-34~\cite{he2016deep}+FFC &21.8M &3.67 &\textbf{73.96\%}
&ResNeXt-50~\cite{xie2017aggregated}(32x4d)+FFC &25.04M &3.78 &\textbf{77.56\%}\\
ResNet-50~\cite{he2016deep} &25.56M  &3.86 &75.85\%
&ResNeXt-101~\cite{xie2017aggregated}(32x4d) &44.18M &7.50 &78.46\% \\
ResNet-50~\cite{he2016deep}+FFC &25.57M & 3.87 &\textbf{76.80\%}
&ResNeXt-101~\cite{xie2017aggregated}(32x4d)+FFC &44.19M &7.51 &\textbf{78.61\%}\\
\hline
\end{tabular}
}
\end{center}
\vspace{-4mm}
\begin{tablenotes}
\scriptsize
\item\hspace*{\fill}\textbf{*} all results are reproduced in the PyTorch framework. 
\end{tablenotes}
\vspace{2mm}
\caption{\textbf{Classification results on ImageNet-1K.} The left side of the table shows the performance of the base that has been reproduced and the right side shows the performance of attaching FFC to the base. }
\label{table:Classification }
\vspace{-5mm}
\end{table}
\vspace{2mm}
\jc{Experiment results and overheads are shown in \tabref{table:Classification }.
FFC consistently shows performance improvements in all backbone architectures, but has negligible overheads.}
\jc{
Among variants of ResNet~\cite{he2016deep}, performance improvements for ResNet-18 and ResNet-34 are relatively small.
We conjecture the small improvements are due to the smaller final features in these architectures. The final feature size of ResNet-18 and 34 are 512, and there is not much room for feature filtering.}
\subsection{Bag-of-Trick, Attention Module}
\label{sec:Bag-of-trick, Attention module}
\jc{We apply FFC upon the techniques introduced in \textit{bag-of-tricks}~\cite{he2019bag}, such as cosine learning rate decay and mixup, to see if FFC can further improve the performance. In addition to the \textit{tricks}, we add attention modules~\cite{hu2018squeeze,lee2019srm} to the baselines, because they also efficiently improve the overall performances. All experiment settings are identical to Sec.~\ref{sec:ImageNet}}
\jc{The results are shown in \tabref{table:bag-of-tricks }. }
\jc{We can observe that FFC can significantly improve upon the tricks and attention modules. The result emphasizes the complementary utility of FFC.}

\begin{table}[!htbp]
\small
\setlength{\tabcolsep}{3pt}
\begin{center}
\resizebox{1.0\textwidth}{!}{
\begin{tabular}{ l|c|c|c||l|c|c|c}
\hline
Refinements & GFLOPs & Top-1 & Top-5 & Architecture & Params & GFLOPs & Top-1\\
\hline
\hline
Efficient(reproduced)~\cite{he2019bag} &4.3 &77.06\% &93.53\%
&ResNet-50~\cite{he2016deep} &25.56M &3.86 &75.85\%\\
Efficient~\cite{he2019bag}$\dagger$ &4.3   &77.16\% &93.52\%
&ResNet-50~\cite{he2016deep}+FFC &25.56M &3.87 &\textbf{76.80\%} \\
+ cosine decay~\cite{he2019bag}$\dagger$ &4.3 &77.91\% &93.81\%
&ResNet-50~\cite{he2016deep}+SE~\cite{hu2018squeeze} &28.09M   &3.87 &76.80\%\\
+ label smoothing~\cite{he2019bag}$\dagger$ &4.3 &75.60\% &94.09\%
&ResNet-50~\cite{he2016deep}+SE~\cite{hu2018squeeze}+FFC &28.09M &3.88 &\textbf{77.54\%}\\
+ mixup w/o distill~\cite{he2019bag}$\dagger$ &4.3 &79.15\% &94.58\%
&ResNet-50~\cite{he2016deep}+SRM~\cite{lee2019srm} &25.62M &3.88 &77.01\%\\
+ FFC &4.3 &\textbf{79.62}\% &\textbf{94.75\%}
&ResNet-50~\cite{he2016deep}+SRM~\cite{lee2019srm}+FFC &25.62M &3.89 &\textbf{77.91\%}\\
\hline
\end{tabular}
}
\begin{tablenotes}
\scriptsize
\item\hspace*{\fill}\textbf{*} All results are reproduced in the PyTorch framework, excluding the ones marked by $\dagger$. 
\end{tablenotes}
\end{center}
\caption{\textbf{The experiment result when FFC is used with} bag-of-tricks and attention modules. The left rows are the experiment results with bag-of-tricks, and the right ones are the results with attention modules. We empirically show that FFC has complementary effect upon bag-of-tricks, attention modules. ‘$\dagger$’ denotes results reported in the original paper~\cite{he2019bag}.}
\label{table:bag-of-tricks }
\end{table}

\subsection{MS COCO detection}
\label{sec:COCO}
\begin{table}[!htbp]
\small
\setlength{\tabcolsep}{3pt}
\begin{center}
\resizebox{0.7\textwidth}{!}{
\begin{tabular}{ l|c|c|c|c}
\hline
Backbone & Detector & mAP@.5 & mAP@.75 & mAP@[.5, .95] \\
\hline
\hline
ResNet-50~\cite{he2016deep} & faster rcnn~\cite{ren2015faster}  & 46.3 &28.0 & 27.1 \\ 
ResNet-50~\cite{he2016deep} & faster rcnn~\cite{ren2015faster}+FFC  & \textbf{47.1} &\textbf{28.6} & \textbf{27.7} \\ 
ResNet-101~\cite{he2016deep} & faster rcnn~\cite{ren2015faster}  & 48.4 &30.7 & 29.1 \\
ResNet-101~\cite{he2016deep} & faster rcnn~\cite{ren2015faster}+FFC & \textbf{49.1} &\textbf{31.4} & \textbf{29.5} \\ 
\hline
\end{tabular}
}
\end{center}
\vspace{2mm}
\caption{\textbf{MS-COCO detection task.} Results of evaluation of FFC using faster rcnn in MS COCO detection task.}
\label{table:detection }
\vspace{-5mm}
\end{table}
\vspace{2mm}
\jctwo{
We choose the Microsoft COCO~\cite{lin2014microsoft} dataset to verify if FFC can improve the performance of object detection. The dataset has 3 splits: train split with 80k images, validation split with 40k images, and test split with 5k images. Following the protocol in \cite{liu2016ssd}, we used train+val splits for training, and use the test split for evaluation.
}
The average mAP over different IoU thresholds from 0.5 to 0.95 is used for evaluation. We adopt Faster-RCNN as our detection method and ImageNet-1K pretrained ResNet-50 and ResNet-101 as our baseline networks.
The result is summarized in \tabref{table:detection }.
FFC shows significant performance improvements in object detection by simply modifying the Classifier to FFC.

\subsection{Cityscapes semantic segmentation } \label{sec:Cityscapes}
\begin{table}[!htbp]
\small
\setlength{\tabcolsep}{3pt}
\begin{center}
\resizebox{1.0\textwidth}{!}{
\begin{tabular}{ l|c|c|c||l|c|c|c}
\hline
Architecture & Input size & Batch size & mIoU & Architecture & Input size & Batch size & mIoU \\
\hline
\hline
PSPNet~\cite{zhao2017pyramid} &768 &8 &78.3 &DeeplabV3~\cite{chen2017rethinking} &768 &8 &78.9 \\
PSPNet~\cite{zhao2017pyramid}+FFC &768 &8 &\textbf{78.8} &DeeplabV3~\cite{chen2017rethinking}+FFC &768 &8
&\textbf{79.3} \\
\hline
\end{tabular}
}
\end{center}
\vspace{-4mm}
\begin{tablenotes}
\scriptsize
\item\hspace*{\fill}\textbf{*} all results are reproduced in the PyTorch framework. 
\end{tablenotes}
\vspace{2mm}
\caption{\textbf{Fine cityscapes semantic segmentation experiment results.} The left is the performance of PSPNet and DeeplabV3, and the right is the performance with FFC attached.}
\label{table:segmentation }
\vspace{-5mm}
\end{table}
\vspace{2mm}
We evaluated FFC in Cityscapes dataset, one of the widely-used benchmark datasets for semantic segmentation task. The dataset with 5,000 fine annotations and 19 labels. 
\jctwo{
We use the open-sourced project segmentation-toolbox~\cite{huang2018pytorch} for the experiments. 
We strictly follow the baseline hyper-parameters in the segmentation-toolbox.
}
We also used the ResNet-101 pretrained weight provided by segmentation-toolbox.
\jctwo{
Please note that we did not use the weights of ResNet-101 + FFC pre-trained on ImageNet-1K, because in this experiment, we want to strictly observe the effect of FFC on segmentation.
}
We simply modified the classifier to FFC for both PSPNet and DeeplabV3. As shown in ~\tabref{table:segmentation }, we showed that our FFC method helps improve performance in segmentation.

\subsection{HMDB51 action recognition in video} \label{sec:HMDB51}
In the case of Action Recognition, it is difficult to configure large datasets such as ImageNet, so pretrained weights in kinetics~\cite{kay2017kinetics} dataset are essential, and it takes 2 weeks to utilize V100x16 in case of I3D~\cite{carreira2017quo}. If pretrained weights cannot be configured on their own due to these problems, it is impossible to change the CNN architecture and a technique that can stably improve performance in fine tuning is also required. 

We evaluated FFC in split 1 of HMDB51, one of the action recognition benchmark datasets. HMDB51 is composed of 51 action categories, and the total number of clips included is approximately 7,000. We used 3D-ResNeXt-101~\cite{hara2018can} for evaluation, and used not only RGB images, but also optical flow extracted by TV-L1~\cite{perez2013tv} algorithm for evaluation. All our experimental environments follow MARS~\cite{crasto2019mars}. Also, in order to prove that FFC is capable of not only training from scratch, but also fine-tuning, the model pretrained from kinetics400 was loaded and fine-tuned. As shown in \tabref{table:ACTION }, FFC showed improved accuracy in both the RGB and optical flow data types in the action recognition task. FFC also showed that not only training from scratch, but also finetuning can maintain the benefits of FFC.
\begin{table}[!h]
\small
\setlength{\tabcolsep}{3pt}
\begin{center}
\resizebox{1.0\textwidth}{!}{
\begin{tabular}{ l|c|c||l|c|c}
\hline
Architecture & Modality & Top-1 & Architecture & Modality &Top-1 \\
\hline
\hline
ResNeXt-101~\cite{hara2018can} &RGB &73.00\%
&ResNeXt-101~\cite{hara2018can}+FFC &RGB &\textbf{74.82\%} \\
ResNeXt-101~\cite{hara2018can}   &Flow &75.90\%
 &ResNeXt-101~\cite{hara2018can}+FFC   &Flow &\textbf{77.45\%}  \\
ResNeXt-101~\cite{hara2018can} &RGB+Flow+MARS~\cite{crasto2019mars} &81.30\% 
&ResNeXt-101~\cite{hara2018can}+FFC &RGB+Flow+MARS~\cite{crasto2019mars} &\textbf{83.46\%} \\
\hline
\end{tabular}
}
\end{center}
\vspace{-4mm}
\begin{tablenotes}
\scriptsize
\item\hspace*{\fill}\textbf{*} all results are reproduced in the PyTorch framework. 
\end{tablenotes}
\vspace{2mm}
\caption{\textbf{3D ResNeXt 101 performance in HMDB51 action recognition benchmark.} 3 channels of RGB modality and 2 channels of optical flow are input. Both modalities stacked 64 consecutive frames, and the spatial size of the image is (112,112). All performance is the result of fine-tuning the pre-trained weight of Kinetics400.}
\label{table:ACTION }
\vspace{-5mm}
\end{table}
\subsection{Ensemble with FFC} \label{sec:ensemble}
\begin{table}[!h]
\small
\setlength{\tabcolsep}{3pt}
\begin{center}
\resizebox{1.0\textwidth}{!}{
\begin{tabular}{ l|c|c|c|c}
\hline
Architecture & Dataset & Epoch & Top-1 Accs & Ensembled Acc \\
\hline
\hline
ResNet-50 & ImageNet-1K &  90 & 75.85\%, 75.61\% & 77.25\%+(1.4\%) \\
ResNet-50+FFC & ImageNet-1K & 90 & 76.80\%, 76.54\% & \textbf{78.32\%+(1.5\%)} \\
\hline
\end{tabular}
}
\end{center}
\vspace{-4mm}
\begin{tablenotes}
\scriptsize
\item\hspace*{\fill}\textbf{*} all results are reproduced in the PyTorch framework. 
\end{tablenotes}
\vspace{2mm}
\caption{\textbf{As a result of experimenting the Ensemble effect of FFC with ResNet-50 architecture in ImageNet-1K.}}
\label{table:Ensemble }
\end{table}
Our FFC is similar to ensemble in that it has multiple outputs and uses them all. For this reason, we experimented with the ResNet-50 model in ImageNet-1K to verify that the effect persists even if we ensemble two models trained with FFC. All experimental settings are exactly the same as in \sref{sec:ImageNet}. As shown in \tabref{table:Ensemble }, our FFC also improves performance in Ensemble. Therefore, our FFC is complementary to the Ensemble method.

\subsection{FFC Visualization with Grad-CAM} \label{sec:grad-cam}
\begin{figure*}[!ht]
  \centering
  \includegraphics[width=\linewidth]{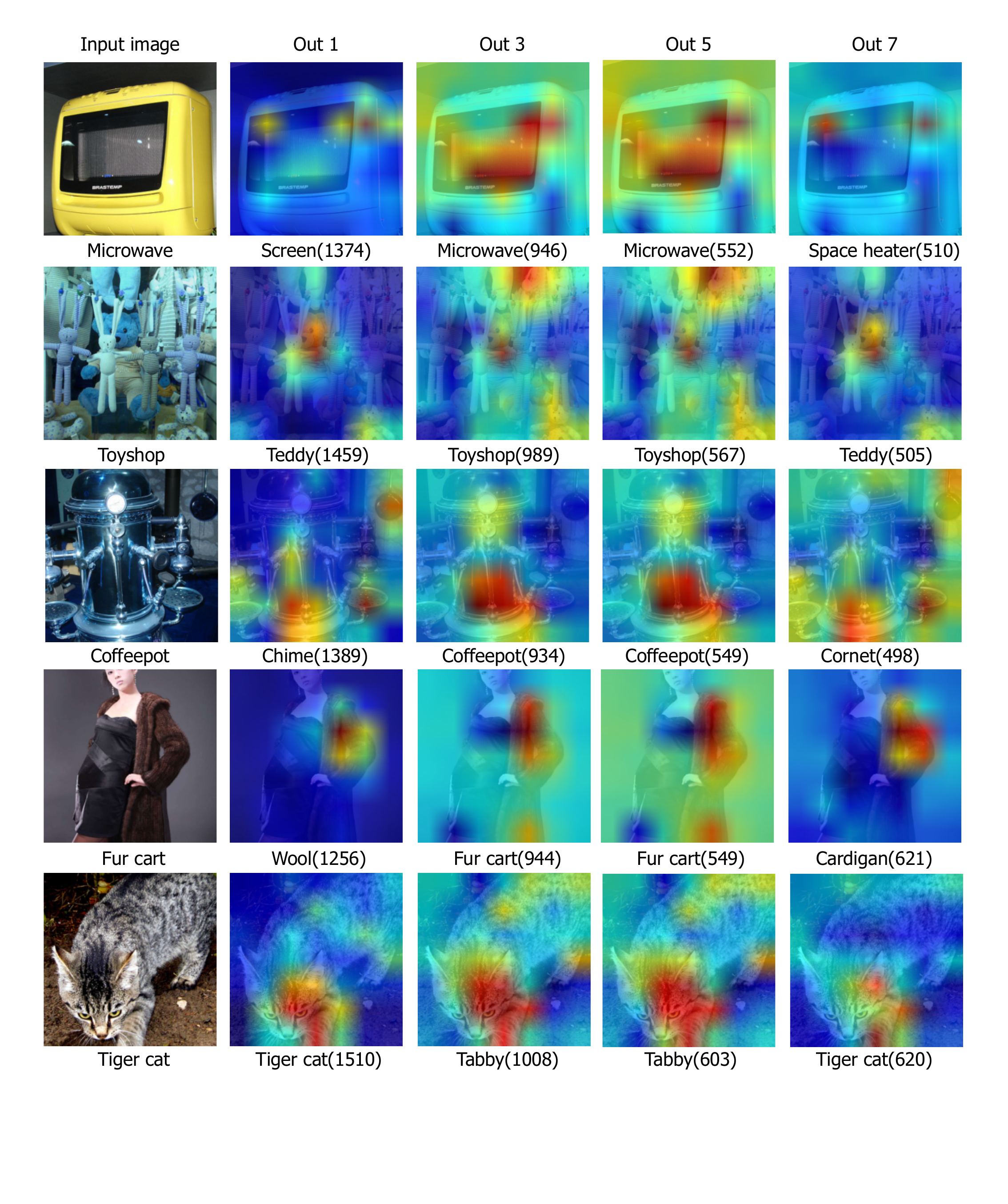}
  \vspace{-18mm}
  \caption{\textbf{Grad-CAM visualization results.} We visualized FFC trained with ResNet-50 architecture in ImageNet-1K. All outputs are output from the shared classifier. As shown in the figure, even if the predictions are all output from the same classifier, their values may be different, and the activated part is also different.`()' is the number of activated units.}
  \label{fig:grad_cam}
\end{figure*}

We adopt Grad-CAM for qualitative analysis of FFC. We use the ResNet-50 + FFC model trained in ImageNet-1K. Also, instead of visualizing all the outputs, Out1, Out3, Out5, and Out7 are visualized.
The differences among the filtered features (Out1/3/5/7) are clearer than the added ones (Out2/4/6), so we can easily analyze by comparing the Grad-CAM visualizations.
As shown in \figref{fig:grad_cam}, multiple features from sequential feature filtering are inputted into a shared classifier and produce different outputs.
While the outputs and the Grad-CAM visualizations vary along with feature filtering, we can observe that the argmax predictions vary among confusing categories. For example, the second sample of `Toyshop', Out1 and Out7 focus on the localized area, and the final prediction is `Teddy'; Out3 and Out5 focus on larger area, and the final prediction is correct.
The majority of the multiple outputs seem to be closer to the ground-truth, so we use voting mechanism as the ensemble method.

\section{Conclusion}
\label{sec:Conclusion}
We propose Sequential Feature Filtering Classifier, a novel classifier that can be attached to any CNN for any recognition task. FFC can significantly improve performance with negligible parameter and computational. In addition, FFC has been extensively verified, including image classification, object detection, semantic segmentation, action recognition, attention modules, and augmentation technique. We even visualized and analyzed FFCs through Grad-CAM. Lastly, we used FFC's multiple outputs simply by voting, but we believe that if we study how to select the output that fits the input image from the multiple outputs according to our intuition, we can make a significant performance improvement.
\bibliography{egbib}
\end{document}